\documentclass[11pt,letterpaper]{article}
\usepackage[T1]{fontenc}
\usepackage{amsmath,amssymb}
\usepackage{newtxtext,newtxmath}
\usepackage[letterpaper,margin=0.96in]{geometry}
\usepackage{microtype,graphicx,booktabs,tabularx,array,caption,placeins}
\usepackage[numbers,square,sort&compress]{natbib}
\usepackage{xcolor}
\definecolor{linkblue}{HTML}{17496B}
\usepackage[colorlinks=true,linkcolor=linkblue,citecolor=linkblue,urlcolor=linkblue]{hyperref}
\newcommand{\papertitle}{Situated Memory for Future Agent Decisions}
\newcommand{\papersubtitle}{Roles, Value, and Feedback}
\newcommand{\paperauthors}{Zhibao Chen\textsuperscript{*,1} and Qian Cheng\textsuperscript{2}}
\makeatletter
\newcommand{\ps@preprint}{%
  \ps@plain
  \def\@oddfoot{%
    \rlap{\footnotesize\textsuperscript{*}\href{mailto:maxzhibao@gmail.com}{\nolinkurl{maxzhibao@gmail.com}}}%
    \hfil\normalfont\thepage\hfil}%
}
\makeatother
\hypersetup{pdftitle={\papertitle: \papersubtitle},pdfauthor={Zhibao Chen and Qian Cheng},
            pdfsubject={Role and value priorities in situated memory for future agent decisions},
            pdfkeywords={agent memory, situated action, goal-derived categories, analogical transfer, outcome feedback}}
\newcolumntype{Y}{>{\raggedright\arraybackslash}X}
\newcolumntype{R}{>{\raggedleft\arraybackslash}p{0.13\linewidth}}
\newcommand{\paperfigure}[2]{%
  \begin{figure}[!htbp]
    \centering
    \includegraphics[width=\linewidth]{figures/#1.pdf}
    \caption{#2}
    \label{fig:#1}
  \end{figure}}
\begin{document}
\thispagestyle{preprint}
\noindent\rule{\linewidth}{2.1pt}\par
\vspace{8pt}
\begin{center}
  {\fontsize{22}{25}\selectfont\bfseries\papertitle\par}
  \vspace{3pt}
  {\fontsize{20}{23}\selectfont\bfseries\papersubtitle\par}
\end{center}
\noindent\rule{\linewidth}{0.55pt}\par
\begin{center}
  {\large \paperauthors\par}
  \vspace{4pt}
  {\small \textsuperscript{1}Huatai Securities, Nanjing, China
  \qquad \textsuperscript{2}OneBeget.com\par}
\end{center}
\vspace{-3pt}
\begin{abstract}
A personal agent must decide what to remember before it knows its next task. The same statement can be approval or advice, depending on who has authority to act. In the situated memory model, the writer compares feasible stores by expected future loss and uses observed action outcomes to revise retention. In paired simulations with a fixed byte budget, changes in authority or error costs reversed some retention choices. In one simulation, verification made outcomes comparable but cost more than it saved. In an exploratory direct-JOIN replay on a MemoryArena travel subset, the mean case-level slot score was 73.89\% for the workflow-structure estimator and 42.84\% for rolling recency. The replay used family-level cross-validation, complete offline demand labels, and reference plans from completed rounds. Adding cues from the preceding request reduced the gain. Cross-domain transfer remains open.
\end{abstract}


\section{Introduction}
Someone tells an agent, \textquotedblleft We can move it to Friday.\textquotedblright\ The agent has room to save the exact date or a handoff link. When the speaker can approve the change and the calendar accepts edits, the date lets a later operator reschedule. When the speaker can only advise, the handoff link identifies where to seek approval. A later audit may need the approval trail.

A situation determines what a role and its priorities imply for a particular decision. A lead may approve a change on one project and only advise on another. The same statement then changes from approval to advice. Reserving a work venue and booking a family trip can both require another person's approval. In either activity the agent can ask who must approve, then use evidence about the particular booking to answer that question. A personal agent encounters such changes as it moves across work, home, and shopping.

We call this architecture a \emph{situated memory model}. Before the next task is known, the writer uses available cues to estimate what the agent could do with each feasible store across likely follow-up tasks. A later correction can update the date itself. A failed handoff can help the writer estimate the cost of omitting the link.

Section~\ref{sec:model} defines memory value for a single write and for a policy that writes repeatedly. Sections~\ref{sec:situation} and~\ref{sec:feedback} test whether authority and error costs change which evidence to retain, and whether observing outcomes pays for itself. Section~\ref{sec:limits} assesses evidence selection and use on public dialogue and travel data. The experiments use explicit stores and, where needed, frozen language models.

\subsection{Concepts, roles, and the value of memory}
Barsalou showed how a goal can bring otherwise dissimilar items into one category. Ratneshwar and colleagues found that both personal and situational goals shape category judgments \citep{barsalou1983adhoc,ratneshwar2001goals}. Gentner's structure mapping identifies relations that can carry across domains. Gick and Holyoak showed experimentally how comparing examples can support schema induction and subsequent transfer \citep{gentner1983structure,gick1983schema}. These accounts explain why an earlier episode may come to mind. A further decision for a bounded agent is which condition or source to preserve now so that it can evaluate the analogy later.

An experience can also reach the agent through someone else's account. Blackmore proposed imitation as a mechanism of cultural transmission \citep{blackmore1999meme}. In a serial-retelling experiment, Mesoudi and Whiten found that low-level event details took up a decreasing share of successive accounts \citep{mesoudi2004transmission}. This motivates keeping the origin of a transferable lesson when later decisions depend on its detail or credibility. Two retellings of one observation can make a rule familiar while providing only one underlying observation.

Situated action locates behavior in its social and material setting. The logic of appropriateness connects a recognized situation to the obligations of a role \citep{suchman1987situated,march2009appropriateness}. We use these insights to specify the feasible actions under a particular delegation. In sensorimotor learning, Heald and colleagues model how inference about context controls memory formation, updating, and expression \citep{heald2021contextual}. Mattar and Daw formalize when accessing a memory can improve future choices. Nicholas and Mattar show that people draw on detailed episodic memories when future demands are uncertain at encoding \citep{mattar2018prioritized,nicholas2026episodic}. These results motivate our treatment of uncertainty and the option to retain detail. Here the design variable is the persistent evidence available to a later agent operating under explicit permissions and costs.

Information value and value-directed compression supply the expected-loss criterion \citep{howard1966information,poupart2002compression}. Earlier work on agent memory scores encoding, forgetting, and retrieval with seven factors before the future question is known \citep{chen2026remember}. A companion study holds the loss fixed and varies the observer's background evidence \citep{chen2026posterior}. Here we hold evidence fixed and ask whether a change in authority or error costs can reverse the retention order. The experiments use a reduced four-factor comparator.

\subsection{Experience in agent memory systems}

Context Language Models edit live context \citep{shao2026context}. MemGPT and related systems manage persistent experience \citep{packer2023memgpt,park2023agents,xu2025amem,chhikara2025mem0,fang2026lightmem,tan2025reflective}. EMBER studies evidence retention under a budget. MGPO attributes marginal utility to memory rewrites \citep{li2026ember,tang2026memory}. MemRL learns the utility of episodic retrieval, and MemCon adapts memory operations using task feedback \citep{zhang2026memrl,jiang2026memcon}. We ask when a change in situation reverses a retention choice, whether observed outcomes distinguish the options, and whether those observations are worth their cost. The same decision comparison could apply to the native memory module in MemT if its effects and costs can be measured \citep{li2026memt}.

Learned Relevance Eviction learns from subsequent reuse to predict which records to retain~\citep{lia2026eviction}. Yang and colleagues withhold later tasks from the writer and compare factual recall with and without the same memory~\citep{yang2026taskutility}. Leto preserves access to source details and uses response feedback to adjust what it retrieves. MAGE organizes execution states in a tree and can revise an earlier branch~\citep{chen2026leto,chen2026mage}. Both report chain-level MemoryArena results. Our travel replay compares evidence selection with a fixed single-slot reader. The paired workflow experiments vary effective authority or the costs of outcomes and test whether action feedback can distinguish retention choices.

MemoraX Code records lessons, prerequisites, and checks~\citep{memoraxcode}. Composer adapts retrieved procedures to the current environment and stage. MemDev-Bench finds that relevant history can mislead when agents skip applicability checks, and that summaries can omit useful steps~\citep{tan2026guide,chen2026memdev}. A qualifier may prevent a wrong action, while the procedure may save work even in an unchanged situation. Before the next task is known, the writer must decide which evidence merits storage.

\section{The situated memory model}
\label{sec:model}

\paperfigure{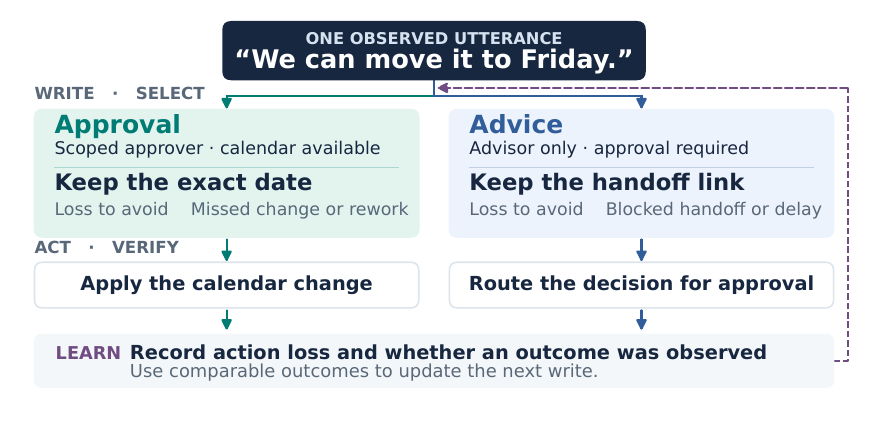}{The same utterance can warrant different memory writes. In the calendar example, scoped authority and tool availability determine which action is feasible. The writer compares the costs of omitting the date or the handoff link under a shared byte cap. At use, the agent checks permissions again. Observed action consequences can update later writes when the outcomes support a comparison.}

\subsection{Represent roles, priorities, and future evidence needs}
At write time \(t\), the agent observes \(o_t\), including the incoming interaction, existing store, and available cues. Represent the relevant situation as \(z_t=(r_t,v_t,w_t)\). The role component \(r_t\) gives responsibilities and authority with their scope. The value component \(v_t\) gives the task's priorities over consequences, for example the relative costs of an erroneous refund and a retry. The workflow component \(w_t\) describes stage and tool availability. A verified permission or a supplied cost profile can be known at writing. A cue to the stage may be uncertain.

Let \(\tau_k\) denote the task at a later decision \(k\), and \(z_k=(r_k,v_k,w_k)\) the situation then. At write time, the writer uses \(o_t\) to predict plausible follow-up tasks and situations, or directly estimates their decision risk. A future audit may need the reference for an inactive arrangement, while execution needs the code for the active one. A role or priority observed at writing can change before either decision.

\paragraph{Transferring an experience.} A decision-relevant category groups episodes by a relation the next task may use, such as requiring approval before making a commitment. One episode can support several such categories. An analogy suggests a relation to check against the new task's authority, scope, and time. In a Bayesian interpretation, applicable evidence updates beliefs about possible consequences. If several interpretations remain plausible, the reader can carry that uncertainty into its prediction or ask for clarification. Task priorities determine how to weigh those consequences and how much evidence to seek. The prediction itself depends on applicable evidence. The experiments supply or extract particular relations. It remains open whether agents can discover new cross-domain analogies from experience.

\subsection{Compare feasible stores by expected decision loss}
Let \(\mathcal F_B(o_t)\) contain the stores that satisfy a persistent byte budget \(B\) and applicable retention constraints. Let \(\pi\) be the shared retrieval, interpretation, and action policy. For a single write followed by one decision, let \(Y_{t+1}^{S,\pi}\) denote the consequences when the writer retains \(S\). The one-decision risk and optimal store are
\begin{equation}
\begin{aligned}
 R_t(S\mid o_t)&=\mathbb E[\ell_{v_{t+1}}(Y_{t+1}^{S,\pi},z_{t+1},\tau_{t+1})\mid o_t],\\
 S_t^\star&\in\arg\min_{S\in\mathcal F_B(o_t)}R_t(S\mid o_t).
\end{aligned}
 \label{eq:write_risk}
\end{equation}
The risk in Equation~\eqref{eq:write_risk} concerns one decision. For decisions through step \(H\), let \(\mu\) govern subsequent writes using the store and available observations at each step. The continuation risk from an initial write \(S\) is
\begin{equation}
 R_{t:H}^{\mu}(S\mid o_t)
 =\mathbb E\!\left[\sum_{k=t+1}^{H}
   \ell_{v_k}(Y_k^{S,\mu,\pi},z_k,\tau_k)\,\middle|\,o_t\right].
 \label{eq:rolling_risk}
\end{equation}
In Equation~\eqref{eq:rolling_risk}, \(Y_k^{S,\mu,\pi}\) depends on the initial store and later writes. A comparison of initial stores fixes \(\mu\). A comparison of rolling writers allows their subsequent stores to differ while matching observations, budgets, and \(\pi\) in the replay. At each decision, \(\pi\) checks the current authority and workflow constraints. \emph{Task value} weights the consequences of permitted actions. \emph{Memory value} is a reduction in the applicable risk when retaining \(S_1\) in place of \(S_0\). An item may add little if it repeats available evidence or become decisive when retained with a qualifier.

The next decision determines the unit of evidence to compare. It could be a quote, a dated field, or a measurement bundled with its calibration. Test the candidate unit against fuller evidence under the same reading conditions. Charge source locators and qualifiers to the budget and disclose separately accounted state. For two options, an error in the risk-gap estimate smaller than the magnitude of the true gap preserves the ordering. Correction can restore the right choice when stale estimates reverse the ranking. A noisy update can also disturb a sound ranking.

\paragraph{When a situational cue changes the write.}
Consider two stores \(S_0,S_1\) that are feasible both before and after observing a cue \(C\). Keep the reader, execution policy, and loss definition fixed. Relative to retaining \(S_0\), the conditional gain from retaining \(S_1\) is \(\Delta(C)=R_t(S_0\mid o_t,C)-R_t(S_1\mid o_t,C)\).
A writer that sees \(C\) takes the better of the two stores and gains \([\Delta(C)]_+\), where \([x]_+=\max(0,x)\). A writer that must commit before seeing \(C\) compares mean risks. By iterated expectation, its best expected gain is \([\mathbb E(\Delta(C)\mid o_t)]_+\). The expected reduction in loss from using the cue is
\begin{equation}
 V(C\mid o_t)=\mathbb E\!\left[[\Delta(C)]_+\mid o_t\right]
   -\left[\mathbb E\!\left[\Delta(C)\mid o_t\right]\right]_+ \ \geq 0.
 \label{eq:cue_value}
\end{equation}
The inequality follows from convexity of the positive-part function. The gain is strict when positive and negative gaps both occur with nonzero conditional probability. If the same store is optimal for every cue value, the gain is zero. In the calendar example, the cue can be whether the delegation covers the project. Equation~\eqref{eq:cue_value} assumes the writer knows the conditional risks. An estimated ranking can lose the gain, and acquiring the cue is worthwhile only if its expected benefit covers its incremental cost.

\paragraph{From action loss to a record score.}
Suppose each decision has one required record. The reader applies it correctly when present, and the decision incurs a fixed loss given the situation when it is absent. Let \(D_j\) indicate that at least one decision requires record \(j\), and let \(c_j\) be the total avoidable loss of decisions that require it. For a store held fixed through the decisions, the reduction in aggregate risk relative to an empty store is
\begin{equation}
 R_t^{\mathrm{fixed}}(\varnothing\mid o_t)-R_t^{\mathrm{fixed}}(S\mid o_t)
   =\sum_{j\in S}\mathbb E[c_jD_j\mid o_t].
 \label{eq:record_value}
\end{equation}
The losses add across decisions, so linearity of expectation suffices even when the \(D_j\) are dependent. With one unit of credit per record used at least once, Equation~\eqref{eq:record_value} yields the probability scores used in the travel packer. These scores approximate Equation~\eqref{eq:rolling_risk} because later writes can evict records and repeated requests can add value. When records substitute for each other or must be used together, the writer compares risks for entire stores.

\subsection{Use action outcomes to revise the comparison}
In Figure~\ref{fig:situated-architecture}, outcomes arrive after action and can refine subsequent writes. The update in Equation~\eqref{eq:feedback_update} estimates which memory choice better serves the specified task priorities. The update takes costs and authorization rules as given. A simple implementation starts from a static estimate \(R_0\). For a discrete evaluation group \(g\) and store \(S\), it computes
\begin{equation}
 \widehat R(g,S)=R_0(g,S)
 +\frac{\sum_{i\in\mathcal I(g,S)}[L_i-R_0(g,S)]}{|\mathcal I(g,S)|+\lambda},
\label{eq:feedback_update}
\end{equation}
Here \(L_i\) is the observed operating loss and \(\mathcal I(g,S)\) includes assigned cases whose outcomes arrived by the cutoff. To derive the update, estimate a correction \(d\) to \(R_0\) by minimizing
\(\sum_{i\in\mathcal I(g,S)}(L_i-R_0-d)^2+\lambda d^2\), with \(\lambda>0\).
The derivative vanishes at \((|\mathcal I(g,S)|+\lambda)d=\sum_i(L_i-R_0)\). Substituting \(R_0+d\) gives Equation~\eqref{eq:feedback_update}. With no observed outcomes the correction is zero. The derivation fits the observed losses. Inferring the target risk additionally requires assumptions about which losses become observable.

Where the candidate policies disagree, check whether the data support a comparison. \textbf{Assignment} concerns which options could be selected and how cases were assigned. Randomization or a justified model of observational assignment can support the comparison. \textbf{Observation} concerns which assigned options yield outcomes by the cutoff. Routine logs can omit labels even after random assignment \citep{lakkaraju2017selective}. \textbf{Precision} concerns whether paired uncertainty estimates can distinguish a meaningful loss difference. When the comparison remains unresolved, the evaluated procedures fall back to the incumbent or acquire further evidence. When costs require interpretation, a learned outcome scorer can help if independent evidence supports its judgments.

\subsection{Account for learning costs and attribute the change}
Subtract the revised policy's operating loss from the incumbent's loss over the same horizon to obtain gross benefit. Subtract incremental observation, exploration, storage, computation, and maintenance costs to estimate net benefit. For \(n_a\) added audits at unit cost \(c\), the audit charge is \(n_a c\). Count an operational retest already included in operating loss once. Report protected error types separately.

Use a shared reader and execution policy to attribute changes to retention. Track whether evidence was nominated, retained, retrieved, and applied. The same store can support different actions if one reader checks a qualifier and the other skips it. The same action score can also come from different interpretations of the situation. Determining the cause can require observed permissions, task labels, or other evidence beyond the final outcome. If the workflow can be replayed, compare alternate stores from the same decision checkpoint, with the same tools and access limits. If comparing an explicit store with a native memory carrier, hold the selected experience fixed and account for both write and read costs \citep[Section 7.3]{li2026memt}.

\section{When the situation changes what to keep}
\label{sec:situation}

\subsection{Task stage and follow-up jointly determine evidence needs}
Two equal-byte records describe an earlier arrangement and a newer proposal. Each includes an execution code and an audit reference. The 84-byte store fits one record. The writer sees uncertain cues to stage and likely follow-up, and the fixed executor receives their realized values after writing. We evaluate 64 held-out synthetic families with the same codec and budget. Four workflows make the best record depend on stage alone, the task alone, their interaction, or neither. The calibrated joint reference uses the stipulated cue probabilities.

\paperfigure{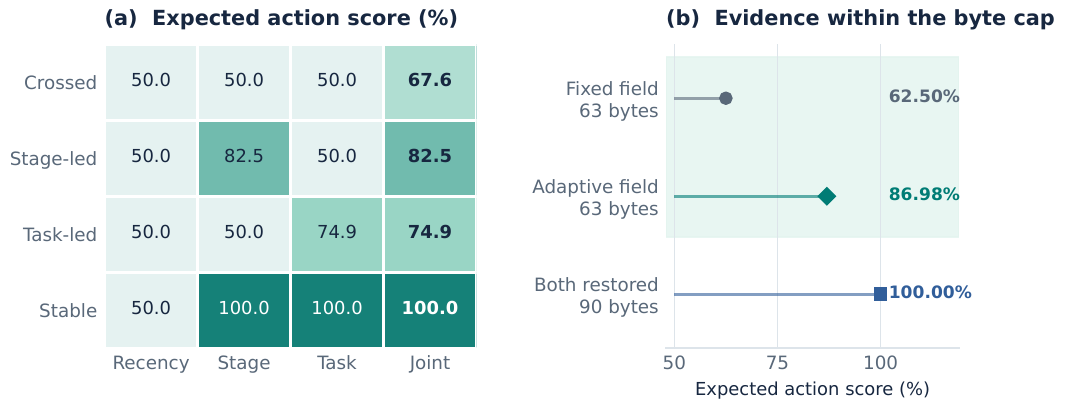}{Joint cues help when both workflow stage and the follow-up task determine the required record. The left panel compares write-time cues across four workflow regimes at 84 bytes. The right panel compares fixed and adaptive field selection at 63 bytes. Restoring both fields at 90 bytes gives the full-evidence reference. Values are expected action scores under the simulation's cue model.}

In the crossed workflow, recency, stage-only, and task-only retention each score 50.00\% in expected action success, versus 67.61\% for the joint reference. Holding either cue fixed and flipping the other reverses the best record in all 32 pairs of each kind. Stable workflows produce no reversals. In a second comparison the record and validation task stay fixed, while the stage determines whether an approval reference or affected scope matters. Adaptive field selection scores 86.98\%, versus 62.50\% for the stronger fixed-field comparator at the same 63-byte cap. A generic sparse record with the same field is byte-identical and yields the same result (Figure~\ref{fig:situation-results}).

The outcome-only writer learned from the success of randomly assigned choices. In the crossed workflow it scored 64.31\%, 62.66\%, and 61.01\% with complete, delayed half, and delayed quarter delivery, respectively. Inverting both stage and task interpretations preserves the observed response distribution for either retention option. Inverting either interpretation alone reverses the preferred option. The writer can therefore improve its choices even when several interpretations fit the same observations. Direct observations of stage or task can resolve the ambiguity.

\subsection{Effective roles change which actions are available}
In the calendar example we hold the wording, speaker, task, and candidate fields fixed. The 36-byte budget fits the date or the link. In 64 synthetic families we vary delegated assignment, project scope, listed title, and calendar availability one at a time. The executor applies the change only with authority for this project and an available calendar. Otherwise handing off is the appropriate next action.

Changing an approver to an advisor, moving the delegation to another project, or making the calendar unavailable switched the preferred evidence from the date to the handoff link in all 64 pairs per contrast. A title change at fixed authorization left the date preferred in every pair. With scoped authority and calendar availability given, the rule selected the correct field on both sides of every pair. The same successful handoff can follow a permissions failure or a system outage.

\subsection{Task priorities change the value of the same evidence}
The calendar study changes what is feasible. The paired refund study changes which consequences matter while holding the notes, eligibility, grants, and execution policy fixed. The writer can retain one note at 224 bytes, about either online or in-store eligibility. Two cost profiles swap the channel where reversal and retry costs are higher. Under identical pre-write evidence, only the writer that ranks notes by action consequences changes its selection across the two profiles in this comparison (243 of 400 pairs, Table~\ref{tab:priority-refund}).

\begin{table}[!htbp]
\centering\small
\caption{Paired cost-profile comparison. Switches compare the same case under the two profiles. Error counts and loss pool the two profiles, with 400 held-out cases each.}
\label{tab:priority-refund}
\begin{tabularx}{\linewidth}{Y R R R R}
\toprule
Retention policy & Switches & Wrong refunds & Rework & Loss\\
\midrule
Four-factor & 0 & 46 & 422 & 1,638\\
Prediction-error feedback & 0 & 24 & 358 & 1,337\\
Action-outcome feedback & 243 & 40 & 404 & 1,249\\
\bottomrule
\end{tabularx}
\end{table}

Pooled across cost profiles, action-outcome feedback incurred 23.7\% less loss than four-factor retention (difference \(-0.486\) per profile-episode, exploratory paired 95\% interval \([-0.654,-0.314]\)). Under in-store priority alone, action-outcome feedback incurred a loss of 827 versus 799 for four-factor retention. Prediction-error feedback had the lowest Brier score (0.173 versus 0.192 for outcome feedback), fewer wrong refunds, and less rework. The difference in weighted loss between the feedback policies remained uncertain. The three measures describe different consequences of the retention choice.

\section{When feedback improves retention}
\label{sec:feedback}

\subsection{Refunds: updating risk within a fixed set of priorities}
We next tested feedback within a fixed cost profile. With delayed feedback (273 of 600 exploration outcomes observed), the action-outcome writer incurred less weighted loss than the prediction-error writer (2{,}916 versus 3{,}066) and issued more wrong refunds (105 versus 100). The known-cost rule had a lower loss point estimate of 2{,}709 and issued 205 wrong refunds. The interval for its loss difference from the action-outcome writer crossed zero. Appendix~\ref{app:feedback-detail} gives the controls.

In the stale-estimate test, the static writer estimates loss from eligibility and note reliability under the shared reader, authorization gate, and executor. The feedback writer uses outcomes from randomized exploration that arrive by the cutoff. The experiment varies starting estimates while keeping the case generator and cost profiles fixed.

When jointly stale estimates reversed the static ranking, feedback changed 10{,}530 of 19{,}200 held-out choices. With delayed feedback at nominal 50\% delivery, weighted loss fell by 38.5\%, wrong refunds by 13.1\%, and rework by 33.6\%. Loss changed by \(-1.077\) per profile-episode (exploratory paired 95\% interval \([-1.107,-1.045]\)). The corrected policy incurred a total loss of 33{,}082, compared with 32{,}793 for the static policy using true parameters. When the starting estimates were correct, feedback increased loss by \(+0.026\) per profile-episode (interval \([+0.005,+0.055]\)). As in the earlier controls, better predictive calibration and lower action loss can favor different choices. Appendix~\ref{app:feedback-detail} reports the controls.

\paperfigure{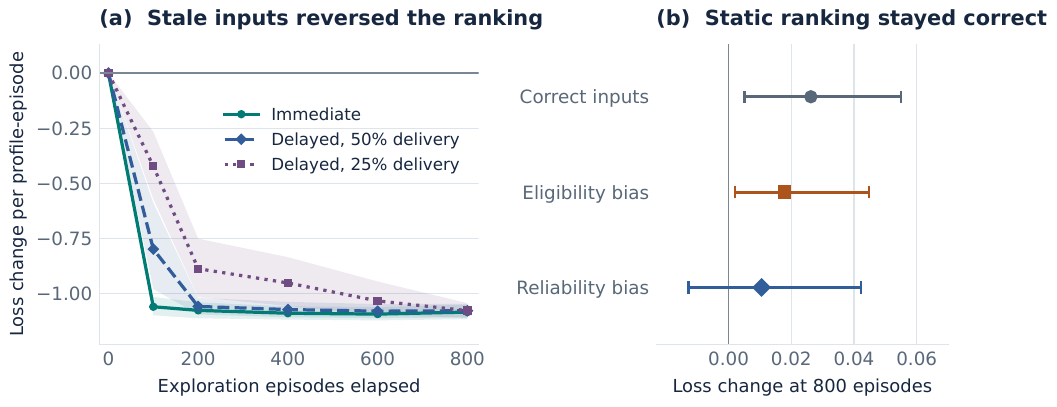}{Feedback reduced loss when stale inputs reversed the original retention ranking. Left, loss relative to static retention across three delivery schedules. Right, the final loss difference at 800 episodes under nominal 50\% delayed feedback when the static ranking stayed correct. Bands and whiskers show exploratory paired 95\% intervals across 12 replicates.}

\subsection{Batch quality control: the value and cost of observing outcomes}
The writer retains measurement and calibration, or inspection and usability, in a 256-byte budget. The executor releases, holds, or retests. Routine logs record conformance after retesting. Even with randomized retention, 12 of 18 situation-option cells have no observed outcomes. The same logs in one release-versus-hold context allow a conformance probability of 0.10 or 0.90. At 0.10 the expected losses of release and hold are 7.2 and 0.3. At 0.90 they are 0.8 and 2.7. The preferred bundle reverses. Even a precise model of the observed retests cannot settle that comparison without an assumption or a new observation about direct decisions.

Random verification added 1{,}494 labels for direct decisions while holding the scored actions fixed. On 5{,}120 held-out batches, operating loss fell 20.2\%, from 6{,}000.8 to 4{,}789.6. The revised policy made 116 fewer false releases, 122 more false holds, and 541 fewer operational retests. At 1.2 units per audit, verification cost 1{,}792.8, leaving a net benefit of \(-581.6\). The break-even price over this horizon was 0.81 units per added audit (Figure~\ref{fig:batch-selective-feedback}).

\paperfigure{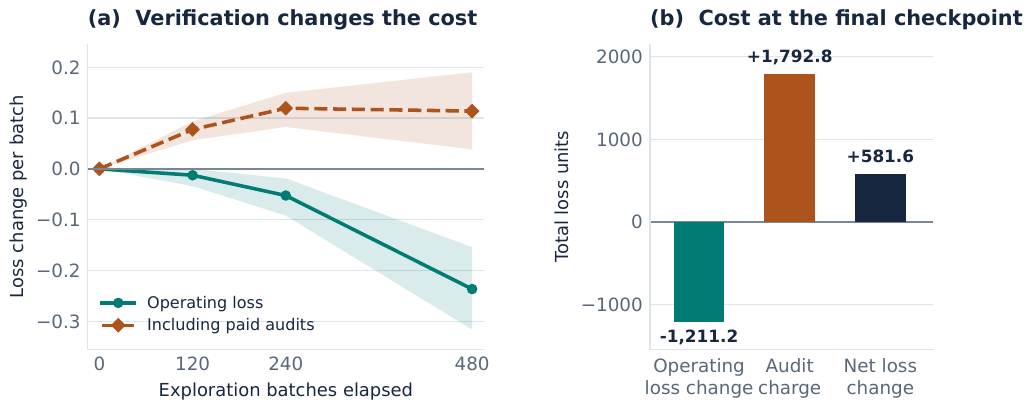}{Verification lowered operating loss but raised total loss after audit charges. Left, changes per batch against the outdated static policy, with exploratory 95\% replicate-bootstrap intervals. Right, the final checkpoint's operating savings, audit charge, and net change in loss over the same 5{,}120 held-out batches.}

\FloatBarrier
\section{Evidence and transfer limits}
\label{sec:limits}

We score individual components of the memory pipeline using tasks from two public benchmarks. For LongMemEval-S, we use the original dialogue and reference answers, judge answers locally, and cap bytes at writing and reading. For MemoryArena, we replay a subset of travel tasks and use the benchmark's single-slot checker. The calendar, refund, and batch comparisons isolate mechanisms in constructed workflows.

\subsection{Extracting facts and using them are separate steps}
On 64 bilingual views from eight calendar families, the frozen Qwen3.5-9B writer selected the correct field in 40 views, Gemma 4 31B in 48, and Qwen3.8 27B MLX in 49. The two larger models extracted all three required facts on every view. When a common rule used their extracted delegation, covered project, and calendar state, both selected the correct field on all 64. The direct and structured procedures changed both the prompt and decision implementation. The comparison supports auditing extraction and application separately.

\subsection{Usable evidence must survive writing and satisfy the task}
Controlled histories separate the effects of retaining more evidence and applying it correctly. Under matched write and read limits, a compact representation preserves both versions in 7 of 36 histories versus 2 with raw retention. Current answers remain at 3 of 36. A reader even returns an expired value in one case where it receives both the closure and the successor. On another 36 histories, holding the compact stores fixed and checking applicability across the whole store raises historical accuracy from 22/36 to 31/36 (paired gain of 25.0 points, exploratory 95\% interval \([11.1,38.9]\)). Current accuracy stays at 6/36. All 12 fresh replacement histories have already lost the new source at writing.

When a replacement is verified, the writer can protect evidence for both current action and later review. Under the same budget and reader, protecting the successor alone raises current replacement accuracy from 2/16 to 13/16 with no change in historical accuracy. Protecting both raises the current count to 15/16 and the historical count from 39/48 to 42/48 (Table~\ref{tab:replacement-policy}).

\begin{table}[!htbp]
\centering\small
\caption{Version protection on 48 constructed histories at 1,450 storage bytes. All policies use the same applicability resolver and question-aligned read limits. Counts concern designated evidence or correct answers.}
\label{tab:replacement-policy}
\begin{tabularx}{\linewidth}{Y R R R R}
\toprule
Writer & Both sources (48) & Current (48) & Historical (48) & Replacement current (16)\\
\midrule
Raw four-factor & 5 & 5 & 35 & 0\\
Compaction & 10 & 12 & 39 & 2\\
Protect successor & 20 & 23 & 39 & 13\\
Protect both & 23 & 25 & 42 & 15\\
\bottomrule
\end{tabularx}
\end{table}

On a 24-question LongMemEval-S subset \citep{wu2025longmemeval}, the four-factor, replacement-only, and three-relation writers answered 8, 7, and 9 questions correctly. We then used unmodified dialogue in two further 24-history holdouts. No replacement passed the first holdout's gate. In the second, claim-first pairing nominated none of the five adjudicated replacements. Raw-turn matching nominated two, both of which the gate rejected. A verbatim quote can still be assigned to the wrong slot or time. In a separate diagnostic, prioritizing annotated evidence from the store repaired none of 24 answers. Giving the reader priority access to annotated evidence from the full history repaired two. Appendix~\ref{app:evidence-detail} traces these failures through nomination, retention, and use.

\subsection{Forecasting evidence demand before the next request}
We screened all 270 released MemoryArena travel rows for direct JOIN clauses that name one prior traveler and have matching source and target reference values~\citep{he2026memoryarena}. After the first three completed requests, 224 rows from 45 itinerary families had eligible clauses, yielding 483 scored slots. Earlier development and case selection drew on the same release. The five outer folds held out families from fitting. At the fixed write boundary, the feedback gate gained \(+0.16\) points over recency by family (exploratory paired 95\% interval \([-0.26,+0.63]\), Figure~\ref{fig:fixed-write-gate}).

At that write boundary, 217 of 483 scored slots required plans from later rounds. Recency missed a further 108 slots whose records fit within the budget. The rolling replay adds each completed reference plan at the end of its round. All policies use the same observations, codec, byte cap per case, and official single-slot checker. A policy's retained records and the newly available plan form the next candidate pool.

In the planned primary comparison, the model using workflow structure and live request cues scored 70.11\%, compared with 42.84\% for rolling recency. The gain by family was \(+27.69\) points (exploratory paired 95\% interval \([+22.16,+33.17]\)). The structural model alone scored 73.89\%. A follow-up baseline with the same offline labels pooled demand by source position and slot and scored 52.44\%. The structural model gained \(+21.74\) points by family over that baseline (interval \([+16.20,+27.32]\)). To isolate selection from the effects of earlier writes, we evaluated all methods on recency's candidate pools. The same-budget one-step hindsight packer filled 391 of 483 slots. Recency filled 211, pooled frequency 268, and the structural model 326. Figure~\ref{fig:application-diagnostics} and Table~\ref{tab:travel-demand} report the rolling results. During training, the learned policies use complete offline demand labels. During scoring, they see each test request only after the corresponding write. All receive the same past reference plans, and the checker scores individual slots.

\paperfigure{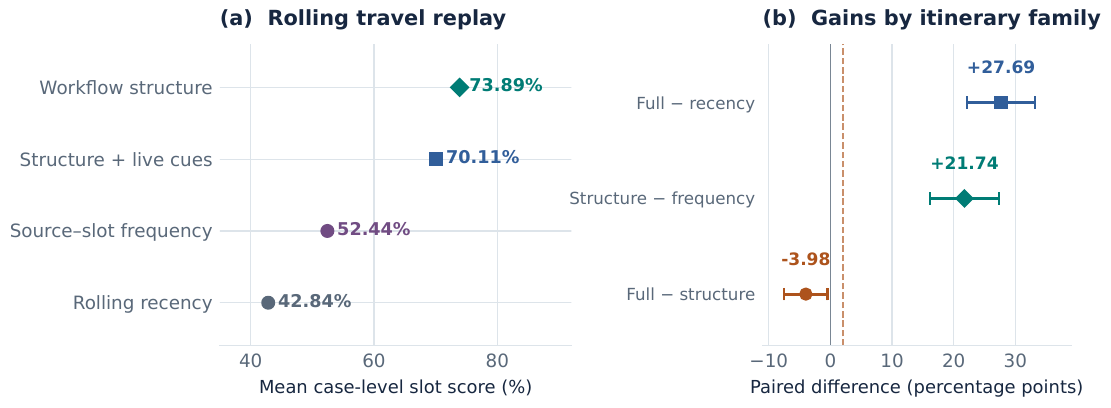}{The workflow-structure model led the rolling travel replay. Left, mean case-level slot scores under the same per-case byte cap. Right, exploratory paired 95\% intervals for family-weighted differences. The full model adds cues from the previous request. The dashed line marks the two-point threshold set for the primary comparison with recency. Table~\ref{tab:travel-demand} reports the counts out of 483.}

\FloatBarrier
\section{Discussion and conclusion}
Evidence demand concerns what a future decision will need to know. Environment forecasts concern what will happen. In the controlled studies, delegated authority and error costs changed the retention value of the same reliable fact. At action time, the reader still had to retrieve the record and check whether it applied. Outcome feedback improved retention choices even when different interpretations of the cues fit the same observations.

Workflow structure improved retention on held-out itineraries from the same benchmark. Adding cues from the preceding request slightly improved record-level Brier loss while filling fewer slots. Under a byte cap, small changes in estimated value can reorder competing records. The replay weights successful slots equally and supplies complete demand labels during training. Role-dependent losses, selective feedback, and transfer require separate evidence.

LongMemEval covers cross-session recall and updates. LoCoMo and LongMemEval-V2 offer further tests through long-conversation questions and event summaries, and through workflow knowledge and premise awareness in web-agent trajectories, respectively \citep{wu2025longmemeval,maharana2024locomo,wu2026longmemevalv2}. A transfer test on unseen families could preserve an approval requirement across domains or change the delegation within one domain. Compare recency, source-domain-only retention, and retention that checks the relation under matched pre-write information, codec, reader, executor, and budgets. Include a no-memory arm to measure absolute utility. Report official scores separately from transfer probes measuring correct reuse, harmful transfer, and subsequent action loss.

Varying situated retention and outcome correction independently would help attribute any gain. Role and priority inference from incomplete traces and longitudinal personal-agent performance remain open. Evidence bytes exclude model parameters and specified auxiliary state. Smaller stores can still require more prompt tokens. Native memory and offline procedure consolidation are compatible extensions \citep{li2026memt}. The tests held observation rules fixed, leaving reflexive changes open.

Roles and workflow state define which actions are available, and task priorities determine how their consequences are valued. Representing both in the situation lets the writer compare the expected losses of feasible stores before the next request arrives. Feedback can revise that comparison when outcomes are sufficiently comparable and the expected savings justify the cost of observing them.

\FloatBarrier
\medskip
\small
\appendix
\section{Experimental protocols and extended results}

\subsection{Situation, authority, and transfer}
\label{app:situation-detail}
\textbf{Record and field selection.} The situation generator uses 64 development and 64 disjoint test families. The crossed workflow requires the live arrangement for execution and the other arrangement for audit. Each family is evaluated at every combination of stage and follow-up cues. Correct supported actions score one, and failures or unsupported actions score zero. The reported scores integrate the known stochastic cue model. In stable or single-dependency controls the target is invariant to one or both cues. Ignoring either cue eliminates the joint advantage in the crossed workflow. Inverting one stage cue lowers aggregate joint performance from 81.24\% to 56.20\%. Field selection uses a fixed sparse JSON codec. Restoring both candidate fields at 90 bytes is an information reference for the 63-byte one-field policy. A fixed reader cannot recover an omitted field under this codec.

The feedback stress test starts with inverted cue calibration and a stationary generator. We randomize choices on development families and deliver either complete feedback or delayed feedback under nested masks. The first learner observes stage, task, and success labels. It recovers the calibrated reference choices at all tested delivery rates. The second observes success alone and fits a score for the selected option by workflow and cue pair using a symmetric one-success, one-failure prior. Stable workflows use their fixed record role. These scores measure decision improvement under partial supervision. The crossed response law makes jointly inverted cues observationally equivalent.

\textbf{Role and authority.} Each of the 64 families has four paired comparisons, sharing a reference view with a scoped approver and an available calendar. All one-field records use 36 serialized bytes. The policies always keep the date, use title, use assignment, use scoped assignment, or combine scoped assignment and calendar availability. Pooled action scores across 512 views are 62.5\%, 50.0\%, 75.0\%, 87.5\%, and 100\%, respectively. A title-only policy solves zero pairs on both sides for every contrast. Assignment alone solves all delegation and title pairs, but misses the scope and availability contrasts. Adding scope solves the scope pairs. Adding tool state also solves the availability pairs. The downstream permission gate is shared, so these scores measure evidence for the appropriate next action, including handoff when required.

\textbf{Frozen model diagnostic.} Eight preselected families supply 64 bilingual prompts. Direct writers receive the rule and choose a field under the same structured response contract. The other procedure asks for assignment, covered project, and calendar availability, then applies a deterministic rule to the returned fields. We request temperature 0.1 and decoding seed 17 and disable thinking. Ollama digests begin \texttt{f6cab3f79f85}, \texttt{6316f0629137}, and \texttt{5642e97495e1}. The first two are Q4\_K\_M, while the MLX replay uses NVFP4. Each model sees the same prompts within a procedure. Qwen3.5-9B extracts all required facts in 57/64 views and selects the correct field in 60/64 after the rule is applied. Gemma's eight direct gains over Qwen3.5-9B concern calendar availability. Qwen3.8 agrees with Gemma on 63/64 direct choices and corrects one delegation case. It still misses 7 delegation cases and 8 project-scope cases in direct choice. The two larger models use different weights, quantization, and backends.

\textbf{MemoryArena.} Five outer folds hold out nine whole itinerary families each. Of the remaining 36 families, eight fit the initial writer, 14 fit feedback updates, and 14 calibrate the gate. All arms use the same compact codec, packing, budget, candidate allocations, and reader. Exploration exposes one assigned choice per case. Initial feedback is complete. Fitting and calibration feedback have nominal half delivery, delays of 2--20 episodes, and a fixed cutoff. The writer proposes a store using pooled source-order and slot estimates. The gate uses 256 resampled sets of calibration families and switches if the tenth percentile of predicted gain is greater than 0.02. Across 24 seeds and 5{,}376 paired evaluations it accepts 381 proposals, including 64 wins, 46 losses, and 271 ties versus recency. The ungated writer's family-weighted gain is \(-0.05\) points (exploratory 95\% interval \([-3.24,+3.21]\)). All families appeared in earlier pilot work. The travel subset evaluates direct JOIN slots and does not cover the benchmark's other task types.

\paperfigure{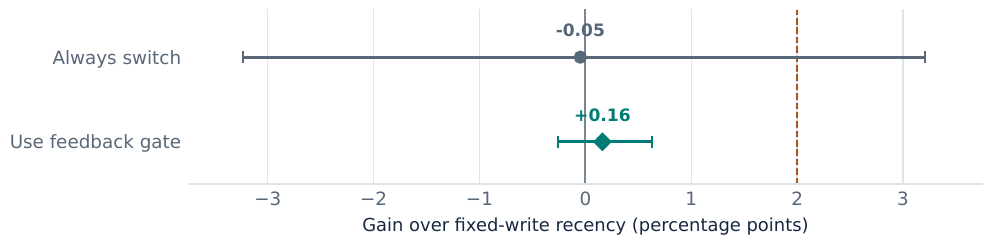}{Fixed-write travel feedback changed little relative to recency. Estimates and exploratory paired 95\% intervals are family-weighted. The dashed line marks the two-point threshold.}

\subsection{Rolling evidence-demand replay}
\label{app:travel-demand}
We use the same 224 cases, five outer folds, and 483 direct JOIN slots. Each write sees the retained store and the latest completed reference plan. The starting write sees the four earlier plans. The agent names known at that round are included in the charged store. An independent evaluator derives whether each record is needed by a remaining scored request. At test time the writer cannot use that label or the next request. A deterministic reader resolves each requested person, day, and slot, and submits the stored value to the original slot checker.

We fit logistic models with an L2 penalty of 0.003 on the non-intercept weights. The packer maximizes the sum of probabilities that each retained record will be needed at least once, subject to the byte cap. This is a surrogate for Equation~\eqref{eq:rolling_risk}. It assigns one unit to a record's future use, even with repeated requests, and does not model retention or application by later writers. Model inputs are categorical indicators. The structural model uses source order, age, elapsed round, day, slot, trip length, and interactions of age with slot and round, trip length with day, and round with slot. The full model adds person mentions in the body of the just-completed request, exact JOIN citations, and equality with the corresponding slot of the new plan. The age-only and age-by-stage models are follow-up ablations. The source--slot baseline uses the same training labels. For each source-position and slot pair, it estimates demand by \((m+0.5)/(n+1)\), where \(m\) is the number of candidate snapshots needed later and \(n\) is the number of candidate snapshots in the training families. All of these measures refer to unit-valued slots on the travel subset.

\begin{table}[!htbp]
\centering\small
\caption{Rolling travel replay on 224 cases and 483 scored slots. Intervals bootstrap paired itinerary families. Scores average per-case fractions. Gains average family differences against rolling recency.}
\label{tab:travel-demand}
\begin{tabularx}{\linewidth}{Y R R >{\raggedleft\arraybackslash}p{0.29\linewidth}}
\toprule
Writer & Slots & Score & Family gain, pp [95\% interval]\\
\midrule
Recency & 211 & 42.84\% & Reference\\
Source--slot frequency & 248 & 52.44\% & \(+9.93\) \([+3.56,+16.04]\)\\
Learned age & 267 & 56.20\% & \(+14.01\) \([+8.42,+19.74]\)\\
Learned age and stage & 269 & 56.75\% & \(+14.30\) \([+8.52,+19.88]\)\\
Workflow structure & 345 & 73.89\% & \(+31.67\) \([+26.05,+37.19]\)\\
Structure and live request cues & 326 & 70.11\% & \(+27.69\) \([+22.16,+33.17]\)\\
\bottomrule
\end{tabularx}
\end{table}

The structural writer improves on recency in 113 cases, ties in 102, and loses in nine. When adding live cues to structure, the family difference is \(-3.98\) points (exploratory paired 95\% interval \([-7.48,-0.47]\)). Brier loss on 37{,}011 record snapshots from recency's candidate pools changes from 0.01897 to 0.01892. The one-step hindsight packer operates on those same pools and uses the next request to maximize the number of correct slots. It fills all 391 available slots within the same byte cap.

\paragraph{Reproducibility.} The travel replay archive in the ancillary material contains the evaluator, frozen family folds, per-case outputs, and commands to obtain and verify the pinned public dataset and scorer. It reproduces Table~\ref{tab:travel-demand} from the released inputs without a model API.

\subsection{Feedback: learning conditions and cost controls}
\label{app:feedback-detail}
\textbf{Refund setup.} Online and in-store eligibility rates are 0.52 and 0.42, with note accuracies 0.90 and 0.77. The reader is frozen Qwen3.5-9B. The executor assigns probabilities 0.82, 0.18, or 0.30 for positive, negative, and missing evidence, respectively. It refunds above 0.5 only with a grant. Reversal costs twice a manual retry; the high-cost channel multiplies both costs by six. Eligibility is observed even after a hold. Each case reports wrong refunds, manual retries, and their combined rework count. Fitted scoring state is outside the note budget.

The first study uses 600 randomized exploration cases and 400 held-out cases per profile. The four-factor score combines emotional intensity, topic relevance, user relevance, and speaker reliability. Its feedback variants add twice the profile-by-option mean of Brier reduction or negative realized cost after normalizing by bounds of two and 21, respectively. The paired difference in loss of the action-outcome writer versus the prediction-error writer is \(-0.110\) per profile-episode (exploratory 95\% interval \([-0.301,+0.078]\)); see Table~\ref{tab:priority-refund}. All arms record zero unauthorized issuances with the same gate.

The second study uses fresh exploration and test cases. Feedback is immediate or delayed by 2--20 episodes, with complete or independently masked delivery. At the main delayed-50\% snapshot, 273 of 600 observations arrive. The known-cost comparator for a positive note uses \(gm(3p-2)\) with posterior eligibility \(p\), grant rate \(g\), and cost multiplier \(m\). A negative note leaves the hold decision unchanged. The consequence writer has lower weighted cost than prediction-error feedback but higher cost than the known-cost benchmark. The known-cost benchmark incurs more wrong refunds (Table~\ref{tab:refund-controls}).

\begin{table}[!htbp]
\centering\small
\caption{Earlier refund controls. Outcomes pool two profiles at the delayed-50\% snapshot.}
\label{tab:refund-controls}
\begin{tabularx}{\linewidth}{Y R R R R}
\toprule
Policy & Wrong refunds & Retries & Rework & Loss\\
\midrule
Four-factor & 138 & 734 & 872 & 3,535\\
Prediction-error feedback & 100 & 676 & 776 & 3,066\\
Action-outcome feedback & 105 & 681 & 786 & 2,916\\
Always keep high-priority note & 133 & 751 & 884 & 2,852\\
Known-cost expected savings & 205 & 464 & 669 & 2,709\\
\bottomrule
\end{tabularx}
\end{table}

The stale-estimate test uses 12 independent replicates, each with 800 randomized exploration cases and 800 held-out cases. Supplied eligibilities and accuracies are correct, biased in eligibility alone (0.28/0.66), biased in reliability alone (0.72/0.93), or jointly stale (eligibility 0.35/0.60, accuracy 0.76/0.90). The loss residual is grouped by profile, note statements, and retained channel, with 12 zero pseudo-observations. Checkpoints are 0, 100, 200, 400, 600, and 800. Intervals pair whole replicates. Under jointly stale estimates with 25\% nominal delivery, the loss difference remains \(-1.076\) per profile-episode (interval \([-1.108,-1.042]\)). The singly biased settings leave the original static ranking unchanged and correction raises aggregate loss point estimates.

\textbf{Batch setup.} Conformance prevalence is 0.68. Actual conditional signal accuracies are 0.91 and 0.73, and retest probabilities are 0.15 and 0.25 independently of conformance. Signals are conditionally independent given conformance. Costs of false release, false hold, and operational retest are 8, 3, and 1.2. Retest yields the correct action. Static misspecified accuracies are 0.59 and 0.91. Sixteen replicates use 480 exploration batches and 320 test batches each. Added verification independently covers 25\% of direct decisions in expectation, with perfect labels and no change to the executed action. A residual update uses eight zero pseudo-observations per recommendation-and-bundle group and falls back to static retention when observation support is absent. One cell remains empty in four audit replicates.

\begin{table}[!htbp]
\centering\small
\caption{Batch-release outcomes over all 5,120 held-out cases. Loss excludes exploratory verification.}
\label{tab:batch}
\begin{tabularx}{\linewidth}{Y R R R R}
\toprule
Policy & False releases & False holds & Retests & Loss\\
\midrule
Outdated static / routine feedback with fallback & 280 & 730 & 1,309 & 6,000.8\\
Randomized verification & 164 & 852 & 768 & 4,789.6\\
Static with true accuracies & 153 & 293 & 430 & 2,619.0\\
\bottomrule
\end{tabularx}
\end{table}

Operating loss under verification is lower by 0.237 per batch (exploratory 95\% interval from \(-0.316\) to \(-0.154\)). With the accumulated verification charge included, incremental cost is \(+0.114\) per batch (interval \([+0.038,+0.190]\)). This comparison prices audits in the shared training stream over the fixed holdout. It does not price alternatives to exploration or changes in real-world auditing practices.

\subsection{Evidence retention, retrieval, and applicability}
\label{app:evidence-detail}
Temporal applicability can require both a source and its effective period. Recording time and validity time can differ \citep{jensen1994temporal,su2026semantic,rasmussen2025zep,song2026tem}. Our diagnostic distinguishes replacement in the same entity, attribute, and scope from a compatible qualification and a value valid in a distinct period. A verified compact record stores a verbatim span or typed value, source locator, observation time, and any valid bounds. A confirmed same-slot replacement can close a prior interval using its own source locator. Those fields compete for the same storage budget.

\textbf{Retention reversals and public-data protocol.} In 24 paired controlled families, a writer directed to inspect source and evidence solved 12 of 24 reversals on a matched template, versus 0 of 24 for lexical deduplication or the unmodified writer. It solved none of eight version-update pairs. An initial template gave lexical deduplication an easy signal (24/24), motivating the lexical control. The public LongMemEval benchmark has 500 questions across five abilities \citep{wu2025longmemeval}. Our LongMemEval-S pilot uses 36 questions to fit four-factor weights and 24 for calibration. A subsequent relation test selects 12 knowledge-update and 12 temporal-reasoning questions, with no reused pilot questions or matching gold evidence. With 32{,}000 storage bytes and an 8{,}000-byte read cap, four-factor scoring retains 67.6\% of annotated evidence, reads 62.1\%, and answers 8/24 correctly. Adding replacement gives 64.2\%, 58.6\%, and 7/24; adding all three relations gives 66.3\%, 60.7\%, and 9/24. MiniLM ranks evidence, Qwen3.5-9B reads, and Gemma 4 31B judges against references. Prioritizing gold turns still in the store leaves accuracy at 8/24. Prioritizing turns from the full history changes it to 10/24, a post-hoc diagnostic adapted from eviction restoration \citep{shen2026eviction}. An earlier seven-factor study reports gold-evidence retention on 479 usable LongMemEval cases with the future question hidden at writing \citep{chen2026remember}. Its retention metric and policy are distinct from the question-answering comparison here.

\textbf{Controlled evidence path.} With matched text and byte caps, applying effective-time labels alone does not improve the slot-check pilot. The raw and labeled arms each answer 2/24 current and 5/24 historical questions. At 1{,}450 storage bytes, compaction preserves both sources in 5/30 histories versus 3/30 under raw retention. In a separate 36-history test with question-aligned read caps, directly rendering compact records preserves both sources in 7/36 cases versus 2/36. Historical accuracy is 25/36 versus 22/36, while current accuracy is 3/36 in each arm. Compact evidence averages 1{,}111 bytes and 801 prompt tokens, against 1{,}383 bytes and 658 tokens for raw evidence. On 36 additional fixed stores, resolving the slot and interval across the whole store raises historical accuracy from 22/36 to 31/36 (exploratory 95\% paired gain from 11.1 to 38.9 points) and leaves current accuracy at 6/36. The resolver admits records only when the requested time falls in their effective interval and abstains if valid values disagree.

Priority eviction on 48 new histories supplies explicit fields and a common reader and resolver. At session boundaries, same-slot proof of a changed value and a valid transition or linked closure determine which records are protected. Unprotected records are evicted first, with four-factor ordering within each class. At 1{,}450 bytes, protecting both versions improves joint retention by 27.1 percentage points over compaction alone (exploratory paired 95\% interval \([14.6,39.6]\)); Table~\ref{tab:replacement-policy} gives the answer counts. The original sensitivity runs at 1{,}100 and 1{,}900 bytes evaluate retention only. Protecting both improves joint retention over compaction across all three budgets.

\textbf{Original dialogue.} Two fresh 24-history holdouts balance knowledge updates and temporal reasoning and exclude repeated evidence and sessions after the question. The first uses at most one cross-session MiniLM nomination per session (threshold 0.55 with shared non-stopword), with Qwen3.5-9B checking slots, relation, and exact spans. The three arms share the same 32{,}000 storage bytes, 12-record and 8{,}000 read-byte limits, and frozen reader and judge. Among 100 pairs no replacement rule activates. Each arm retains 71.25\%, reads 65.42\%, and answers 7/9 current and 3/14 historical questions. In the annotated-pair review one of six adjudicated replacements was nominated and labeled qualification. Forced presentation of 11 pairs yields one true and two false replacement labels, all blocked by the gate. Explicit expiry recall remains untested in that diagnostic.

\paperfigure{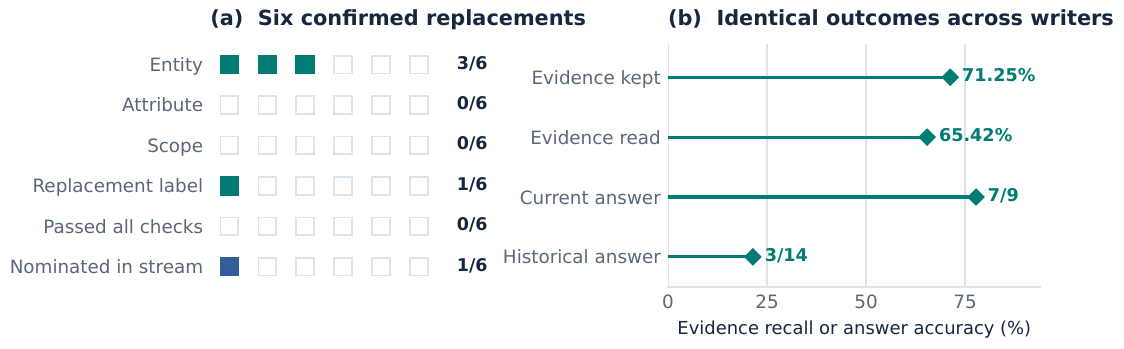}{No confirmed replacement passed the gate on the first unmodified LongMemEval-S holdout. Left, the six confirmed replacements are counted separately for each check. Right, compaction, successor protection, and dual protection attained identical results because the rule never activated. Evidence recall and answer accuracy use the denominators shown.}

The second holdout compares raw nomination with a procedure that extracts source-grounded claims in batches of six turns and then matches slots. The latter ranks with weights of 0.85 for slot similarity and 0.15 for turn similarity. It admits candidates at 0.78, or at 0.64 when they share attribute terms. The budgets and reader match the first holdout. All five adjudicated earlier endpoints are available. Raw pairing nominates two true pairs and accepts none. Claim extraction nominates no true pairs and accepts four false pairs. Both retain 85.42\% of annotated evidence and read 66.67\%. The raw and extraction-first arms answer 6/8 and 7/8 current questions, respectively, and both answer 2/14 historical questions. The additional correct answer followed a presentation change from a false link. Correct version extraction remains unverified. The intervention changes candidate screening, nomination, and the relation prompt together.

\FloatBarrier
\subsection{Auxiliary diagnostic: provenance in state forecasting}
The forecasting study tests a separate question: how source dependence and scope affect updates to a world-state forecast. It uses 24 synthetic families with six rounds, a 512-byte evidence budget, and a frozen extractor. One-route, fixed two-route, and scope-aware two-route aggregators receive the same initial observation. The latter counts each applicable source once. Later rounds contrast repeated references with a second independent draw. A scalar per-scope forecast is carried outside the evidence cap. The recorder maps extracted values to design probabilities 0.15 and 0.85. The update keeps 0.60 of prior log odds, adds 0.40 of the default log odds, and weights the incoming support by 1.25 with per-route weight 0.5.

Under repeated sourcing, source- and scope-aware fusion lowers false updates from 27.1\% to 2.4\% relative to fixed two-route fusion, while persistent changes detected fall from 22/24 to 0/24. Under independent sourcing, false updates fall from 23.6\% to 19.4\%, changes detected fall from 24/24 to 22/24, and Brier loss increases from 0.220 to 0.254. The forecast experiment supplies source and scope labels and measures the trade-off between resisting noise and adapting rapidly (Figure~\ref{fig:multiroute-forecast}).

\newpage
\noindent
\begin{minipage}{\linewidth}
\centering
\includegraphics[width=\linewidth]{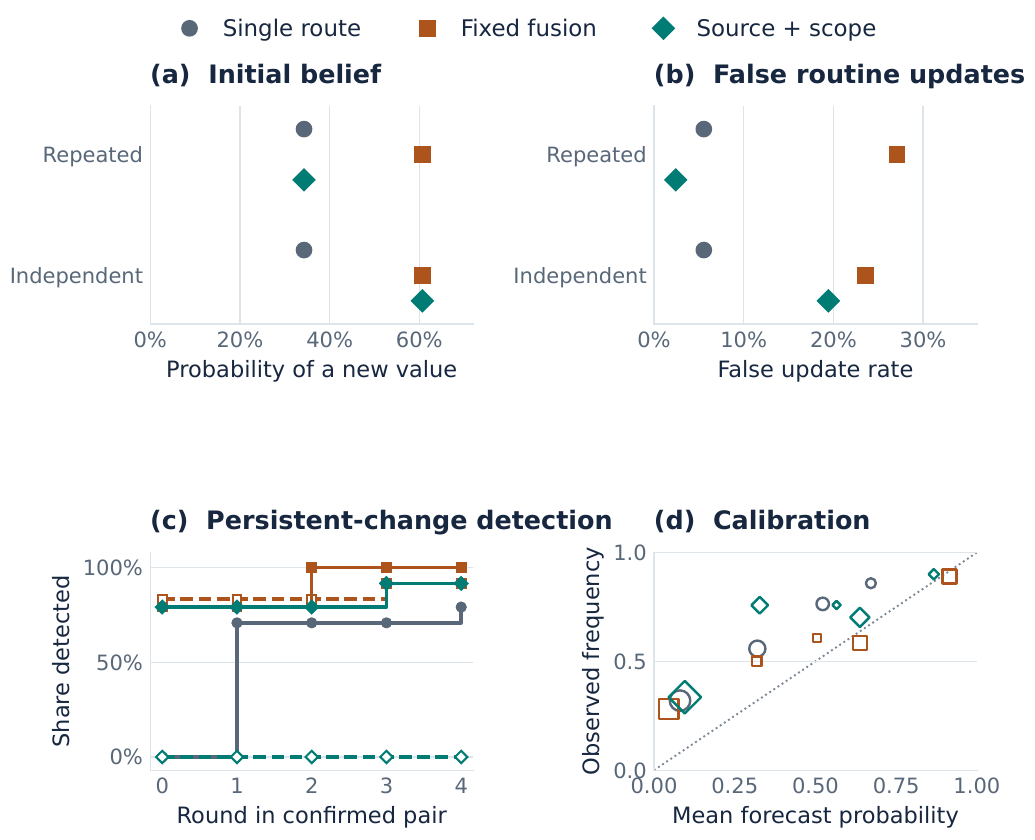}
\captionof{figure}{Auxiliary state-forecast diagnostic under a matched evidence budget. In panel (c), dashed curves indicate repeated references to a single source and solid curves indicate independent sources. Panel (d) uses independent sources. Marker area reflects the count in each nonempty calibration bin, subject to a minimum size for legibility. The dotted diagonal marks perfect calibration.}
\label{fig:multiroute-forecast}
\end{minipage}
\par

\FloatBarrier
\normalsize
\renewcommand{\bibfont}{\footnotesize}
\setlength{\bibsep}{0.3pt plus 0.3pt}
\bibliographystyle{unsrtnat}
\clearpage
\bibliography{references}

\begin{thebibliography}{44}
\providecommand{\natexlab}[1]{#1}
\providecommand{\url}[1]{\texttt{#1}}
\expandafter\ifx\csname urlstyle\endcsname\relax
  \providecommand{\doi}[1]{doi: #1}\else
  \providecommand{\doi}{doi: \begingroup \urlstyle{rm}\Url}\fi

\bibitem[Barsalou(1983)]{barsalou1983adhoc}
Lawrence~W. Barsalou.
\newblock Ad hoc categories.
\newblock \emph{Memory \& Cognition}, 11:\penalty0 211--227, 1983.
\newblock URL \url{https://doi.org/10.3758/BF03196968}.

\bibitem[Ratneshwar et~al.(2001)Ratneshwar, Barsalou, Pechmann, and
  Moore]{ratneshwar2001goals}
S.~Ratneshwar, Lawrence~W. Barsalou, Cornelia Pechmann, and Melissa Moore.
\newblock Goal-derived categories: The role of personal and situational goals
  in category representations.
\newblock \emph{Journal of Consumer Psychology}, 10:\penalty0 147--157, 2001.
\newblock URL \url{https://doi.org/10.1207/S15327663JCP1003_3}.

\bibitem[Gentner(1983)]{gentner1983structure}
Dedre Gentner.
\newblock Structure-mapping: A theoretical framework for analogy.
\newblock \emph{Cognitive Science}, 7:\penalty0 155--170, 1983.
\newblock URL \url{https://doi.org/10.1207/s15516709cog0702_3}.

\bibitem[Gick and Holyoak(1983)]{gick1983schema}
Mary~L. Gick and Keith~J. Holyoak.
\newblock Schema induction and analogical transfer.
\newblock \emph{Cognitive Psychology}, 15:\penalty0 1--38, 1983.
\newblock URL \url{https://doi.org/10.1016/0010-0285(83)90002-6}.

\bibitem[Blackmore(1999)]{blackmore1999meme}
Susan Blackmore.
\newblock \emph{The Meme Machine}.
\newblock Oxford University Press, 1999.
\newblock URL \url{https://www.susanblackmore.uk/the-meme-machine/synopsis/}.

\bibitem[Mesoudi and Whiten(2004)]{mesoudi2004transmission}
Alex Mesoudi and Andrew Whiten.
\newblock The hierarchical transformation of event knowledge in human cultural
  transmission.
\newblock \emph{Journal of Cognition and Culture}, 4:\penalty0 1--24, 2004.
\newblock URL
  \url{https://www.alexmesoudi.com/publication/mesoudi-hierarchical-2004/}.

\bibitem[Suchman(1987)]{suchman1987situated}
Lucy~A. Suchman.
\newblock \emph{Plans and Situated Actions: The Problem of Human-Machine
  Communication}.
\newblock Cambridge University Press, 1987.
\newblock URL
  \url{https://assets.cambridge.org/97805213/37397/frontmatter/9780521337397_frontmatter.pdf}.

\bibitem[March and Olsen(2009)]{march2009appropriateness}
James~G. March and Johan~P. Olsen.
\newblock The logic of appropriateness.
\newblock Oxford University Press, 2009.
\newblock URL \url{https://doi.org/10.1093/oxfordhb/9780199548453.003.0034}.

\bibitem[Heald et~al.(2021)Heald, Lengyel, and Wolpert]{heald2021contextual}
James~B. Heald, M{\'a}t{\'e} Lengyel, and Daniel~M. Wolpert.
\newblock Contextual inference underlies the learning of sensorimotor
  repertoires.
\newblock \emph{Nature}, 600:\penalty0 489--493, 2021.
\newblock URL \url{https://doi.org/10.1038/s41586-021-04129-3}.

\bibitem[Mattar and Daw(2018)]{mattar2018prioritized}
Marcelo~G. Mattar and Nathaniel~D. Daw.
\newblock Prioritized memory access explains planning and hippocampal replay.
\newblock \emph{Nature Neuroscience}, 21:\penalty0 1609--1617, 2018.
\newblock URL \url{https://doi.org/10.1038/s41593-018-0232-z}.

\bibitem[Nicholas and Mattar(2026)]{nicholas2026episodic}
Jonathan Nicholas and Marcelo~G. Mattar.
\newblock Episodic memory facilitates flexible decision-making via access to
  detailed events.
\newblock \emph{Nature Human Behaviour}, 10:\penalty0 760--776, 2026.
\newblock URL \url{https://doi.org/10.1038/s41562-025-02383-3}.

\bibitem[Howard(1966)]{howard1966information}
R.~A. Howard.
\newblock {Information Value Theory}.
\newblock IEEE Transactions on Systems Science and Cybernetics 2(1), 22--26,
  1966.
\newblock URL \url{https://doi.org/10.1109/TSSC.1966.300074}.

\bibitem[Poupart and Boutilier(2002)]{poupart2002compression}
P.~Poupart and C.~Boutilier.
\newblock {Value-Directed Compression of POMDPs}.
\newblock Advances in Neural Information Processing Systems 15, 2002.
\newblock URL
  \url{https://papers.neurips.cc/paper/2192-value-directed-compression-of-pomdps.pdf}.

\bibitem[Chen and Cheng(2026)]{chen2026remember}
Z.~Chen and Q.~Cheng.
\newblock {Learning what to remember: A cognitively grounded multi-factor value
  model for agentic memory}.
\newblock arXiv:2606.12945v2, 2026.
\newblock URL \url{https://arxiv.org/abs/2606.12945v2}.

\bibitem[Chen(2026)]{chen2026posterior}
Zhibao Chen.
\newblock Value-based memory for mental world modeling: Posterior-conditioned
  selection and value-ranking reversals.
\newblock Companion manuscript, version 0.5, 2026.

\bibitem[Shao et~al.(2026)Shao, Shen, Yin, Li, Wang, Ivison, Poovendran,
  Lambert, Xiao, Lewis, Yih, Zettlemoyer, and Koh]{shao2026context}
R.~Shao, S.~Z. Shen, J.~O. Yin, Y.~Li, M.~Wang, H.~Ivison, R.~Poovendran,
  N.~Lambert, T.~Xiao, M.~Lewis, W.~Yih, L.~Zettlemoyer, and P.~W. Koh.
\newblock {Context language models}.
\newblock arXiv:2609.37725v1, 2026.
\newblock URL \url{https://arxiv.org/abs/2609.37725v1}.

\bibitem[Packer et~al.(2023)Packer, Wooders, Lin, Fang, Patil, Stoica, and
  Gonzalez]{packer2023memgpt}
C.~Packer, S.~Wooders, K.~Lin, V.~Fang, S.~G. Patil, I.~Stoica, and J.~E.
  Gonzalez.
\newblock {MemGPT: Towards LLMs as Operating Systems}.
\newblock arXiv:2310.08560v1, 2023.
\newblock URL \url{https://arxiv.org/abs/2310.08560v1}.

\bibitem[Park et~al.(2023)Park, O'Brien, Cai, Morris, Liang, and
  Bernstein]{park2023agents}
J.~S. Park, J.~C. O'Brien, C.~J. Cai, M.~R. Morris, P.~Liang, and M.~S.
  Bernstein.
\newblock {Generative Agents: Interactive Simulacra of Human Behavior}.
\newblock ACM UIST, 2023.
\newblock URL \url{https://doi.org/10.1145/3586183.3606763}.

\bibitem[Xu et~al.(2025)Xu, Liang, Mei, Gao, Tan, and Zhang]{xu2025amem}
W.~Xu, Z.~Liang, K.~Mei, H.~Gao, J.~Tan, and Y.~Zhang.
\newblock {A-MEM: Agentic Memory for LLM Agents}.
\newblock NeurIPS, 2025.
\newblock URL \url{https://arxiv.org/abs/2502.12110v11}.

\bibitem[Chhikara et~al.(2025)Chhikara, Khant, Aryan, Singh, and
  Yadav]{chhikara2025mem0}
P.~Chhikara, D.~Khant, S.~Aryan, T.~Singh, and D.~Yadav.
\newblock {Mem0: Building Production-Ready AI Agents with Scalable Long-Term
  Memory}.
\newblock arXiv:2504.19413, 2025.
\newblock URL \url{https://arxiv.org/abs/2504.19413}.

\bibitem[Fang et~al.(2026)Fang, Deng, Xu, Jiang, Tang, Xu, Deng, Yao, Wang,
  Qiao, Chen, and Zhang]{fang2026lightmem}
J.~Fang, X.~Deng, H.~Xu, Z.~Jiang, Y.~Tang, Z.~Xu, S.~Deng, Y.~Yao, M.~Wang,
  S.~Qiao, H.~Chen, and N.~Zhang.
\newblock {LightMem: Lightweight and Efficient Memory-Augmented Generation}.
\newblock ICLR, 2026.
\newblock URL \url{https://arxiv.org/abs/2510.18866v4}.

\bibitem[Tan et~al.(2025)Tan, Yan, Hsu, Han, Wang, Le, Song, Chen, Palangi,
  Lee, Iyer, Chen, Liu, Lee, and Pfister]{tan2025reflective}
Z.~Tan, J.~Yan, I.-H. Hsu, R.~Han, Z.~Wang, L.~Le, Y.~Song, Y.~Chen,
  H.~Palangi, G.~Lee, A.~R. Iyer, T.~Chen, H.~Liu, C.-Y. Lee, and T.~Pfister.
\newblock {In Prospect and Retrospect: Reflective Memory Management for
  Long-term Personalized Dialogue Agents}.
\newblock ACL, pp. 8416–8439, 2025.
\newblock URL \url{https://aclanthology.org/2025.acl-long.413/}.

\bibitem[Li et~al.(2026{\natexlab{a}})Li, Banerjee, and Che]{li2026ember}
Y.~Li, S.~Banerjee, and T.~Che.
\newblock {EMBER: Efficient memory via budgeted evidence retention for
  long-horizon agents}.
\newblock arXiv:2606.05894v2, 2026{\natexlab{a}}.
\newblock URL \url{https://arxiv.org/abs/2606.05894v2}.

\bibitem[Tang et~al.(2026)Tang, Liu, and Sarabi]{tang2026memory}
J.~Tang, M.~Liu, and A.~Sarabi.
\newblock {Learning what to remember: Long-horizon counterfactual memory
  optimization}.
\newblock arXiv:2609.37930v1, 2026.
\newblock URL \url{https://arxiv.org/abs/2609.37930v1}.

\bibitem[Zhang et~al.(2026)Zhang, Wang, Zhou, et~al.]{zhang2026memrl}
Shengtao Zhang, Jiaqian Wang, Jiayuan Zhou, et~al.
\newblock {MemRL: Self-Evolving Agents via Runtime Reinforcement Learning on
  Episodic Memory}.
\newblock arXiv:2601.03192v2, 2026.
\newblock URL \url{https://arxiv.org/abs/2601.03192v2}.

\bibitem[Jiang et~al.(2026)Jiang, Zhang, Wu, et~al.]{jiang2026memcon}
Eric~Hanchen Jiang, Zhi Zhang, Yuchen Wu, et~al.
\newblock {Memory as a Controlled Process: Learned Adaptive Memory Management
  for LLM Agents}.
\newblock arXiv:2607.13591v3, 2026.
\newblock URL \url{https://arxiv.org/abs/2607.13591v3}.

\bibitem[Li et~al.(2026{\natexlab{b}})Li, Zhang, Liu, et~al.]{li2026memt}
Z.~Li, Z.~Zhang, L.~Liu, et~al.
\newblock {MemT-1.0: Towards Memory-Native Evolution}.
\newblock arXiv:2610.26825v4, 2026{\natexlab{b}}.
\newblock URL \url{https://arxiv.org/abs/2610.26825v4}.

\bibitem[Lia and Mazumder(2026)]{lia2026eviction}
Nusrat~Jahan Lia and Aritra Mazumder.
\newblock {Learning What Not to Forget: Long-Horizon Agent Memory from a Few
  Kilobytes of Learning}.
\newblock arXiv:2606.20954, 2026.
\newblock URL \url{https://arxiv.org/abs/2606.20954}.

\bibitem[Yang et~al.(2026)Yang, Chowdhury, Yan, and
  Thakur]{yang2026taskutility}
Chih-Hsuan Yang, Anjir~Ahmed Chowdhury, Feng Yan, and Rajeev Thakur.
\newblock {What Is This Memory Useful For? Measuring Task-Conditioned Utility
  in Long-Term Agent Memory}.
\newblock arXiv:2611.03930, 2026.
\newblock URL \url{https://arxiv.org/abs/2611.03930}.

\bibitem[Chen and Paracha(2026)]{chen2026leto}
Shuguang Chen and Salman Paracha.
\newblock {Leto: Utility-Centric Agent Memory}.
\newblock arXiv:2611.33005, 2026.
\newblock URL \url{https://arxiv.org/abs/2611.33005}.

\bibitem[Chen et~al.(2026{\natexlab{a}})Chen, Lai, Feng, Han, Zhang, Lu, Li,
  Wang, Jin, Wu, Li, and Chen]{chen2026mage}
Yaoqi Chen, Haibin Lai, Yuru Feng, Chuyu Han, Qianxi Zhang, Baotong Lu, Menghao
  Li, Xinjiang Wang, Zewen Jin, Hao Wu, Cheng Li, and Qi~Chen.
\newblock {Beyond Semantic Organization: Memory as Execution State Management
  for Long-Horizon Agents}.
\newblock arXiv:2606.06090, 2026{\natexlab{a}}.
\newblock URL \url{https://arxiv.org/abs/2606.06090}.

\bibitem[{MemoraX AI}(2026)]{memoraxcode}
{MemoraX AI}.
\newblock {MemoraX Code}.
\newblock Project documentation, 2026.
\newblock URL \url{https://github.com/memorax-ai/memorax-code}.

\bibitem[Tan et~al.(2026)Tan, Gu, Li, Hu, He, Li, and Hao]{tan2026guide}
Zhixuan Tan, Pengjie Gu, Zhao Li, Yihan Hu, Xu~He, Dong Li, and Jianye Hao.
\newblock From memory to guide: Spatio-temporal composer for procedural coding
  memory.
\newblock arXiv:2610.04868, 2026.
\newblock URL \url{https://arxiv.org/abs/2610.04868}.

\bibitem[Chen et~al.(2026{\natexlab{b}})Chen, Liu, Huang, Zhang, Jiang, Wang,
  Sun, He, Li, Hu, Hao, and Feng]{chen2026memdev}
Guorui Chen, Tao Liu, Zhenzhen Huang, Yidan Zhang, Liyu Jiang, Shuoyuan Wang,
  Hanchen Sun, Xu~He, Dong Li, Yihan Hu, Jianye Hao, and Lei Feng.
\newblock {MemDev-Bench}: Does memory help code agents? benchmarking memory
  systems over real repository histories.
\newblock arXiv:2610.27524, 2026{\natexlab{b}}.
\newblock URL \url{https://arxiv.org/abs/2610.27524}.

\bibitem[Lakkaraju et~al.(2017)Lakkaraju, Kleinberg, Leskovec, Ludwig, and
  Mullainathan]{lakkaraju2017selective}
H.~Lakkaraju, J.~Kleinberg, J.~Leskovec, J.~Ludwig, and S.~Mullainathan.
\newblock {The Selective Labels Problem: Evaluating Algorithmic Predictions in
  the Presence of Unobservables}.
\newblock ACM SIGKDD, 275--284, 2017.
\newblock URL \url{https://doi.org/10.1145/3097983.3098066}.

\bibitem[Wu et~al.(2025)Wu, Wang, Yu, Zhang, Chang, and Yu]{wu2025longmemeval}
D.~Wu, H.~Wang, W.~Yu, Y.~Zhang, K.-W. Chang, and D.~Yu.
\newblock {LongMemEval: Benchmarking chat assistants on long-term interactive
  memory}.
\newblock arXiv:2410.10813v2, 2025.
\newblock URL \url{https://arxiv.org/abs/2410.10813v2}.

\bibitem[He et~al.(2026)He, Wang, Zhi, et~al.]{he2026memoryarena}
Zexue He, Yu~Wang, Churan Zhi, et~al.
\newblock {MemoryArena: Benchmarking Agent Memory in Interdependent
  Multi-Session Agentic Tasks}.
\newblock arXiv:2602.16313, 2026.
\newblock URL \url{https://arxiv.org/abs/2602.16313}.

\bibitem[Maharana et~al.(2024)Maharana, Lee, Tulyakov, Bansal, Barbieri, and
  Fang]{maharana2024locomo}
A.~Maharana, D.-H. Lee, S.~Tulyakov, M.~Bansal, F.~Barbieri, and Y.~Fang.
\newblock {Evaluating Very Long-Term Conversational Memory of LLM Agents}.
\newblock ACL, pp. 13851–13870, 2024.
\newblock URL \url{https://aclanthology.org/2024.acl-long.747/}.

\bibitem[Wu et~al.(2026)Wu, Ji, Kawatkar, Kwan, Gu, Peng, and
  Chang]{wu2026longmemevalv2}
Di~Wu, Zixiang Ji, Asmi Kawatkar, Bryan Kwan, Jia-Chen Gu, Nanyun Peng, and
  Kai-Wei Chang.
\newblock {LongMemEval-V2}: Evaluating long-term agent memory toward
  experienced colleagues.
\newblock arXiv:2605.12493, 2026.
\newblock URL \url{https://arxiv.org/abs/2605.12493}.

\bibitem[Jensen and Snodgrass(1994)]{jensen1994temporal}
C.~S. Jensen and R.~Snodgrass.
\newblock {Temporal Specialization and Generalization}.
\newblock IEEE Transactions on Knowledge and Data Engineering, 6(6), 954–974,
  1994.
\newblock URL \url{https://www2.cs.arizona.edu/~rts/pubs/TKDEDec94.pdf}.

\bibitem[Su et~al.(2026)Su, Guo, Hou, Bai, Li, Zhang, Yin, Lin, Jin, Guo, and
  Cheng]{su2026semantic}
M.~Su, Y.~Guo, Z.~Hou, L.~Bai, Z.~Li, Y.~Zhang, G.~Yin, W.~Lin, X.~Jin, J.~Guo,
  and X.~Cheng.
\newblock {Beyond dialogue time: Temporal semantic memory for personalized LLM
  agents}.
\newblock arXiv:2601.07468v2, 2026.
\newblock URL \url{https://arxiv.org/abs/2601.07468v2}.

\bibitem[Rasmussen et~al.(2025)Rasmussen, Paliychuk, Beauvais, Ryan, and
  Chalef]{rasmussen2025zep}
P.~Rasmussen, P.~Paliychuk, T.~Beauvais, J.~Ryan, and D.~Chalef.
\newblock {Zep: A temporal knowledge graph architecture for agent memory}.
\newblock arXiv:2501.13956v1, 2025.
\newblock URL \url{https://arxiv.org/abs/2501.13956v1}.

\bibitem[Song et~al.(2026)Song, Zhuang, and Shi]{song2026tem}
P.~Song, Y.~Zhuang, and L.~Shi.
\newblock {TEM: Temporal state tracking memory for large language model
  agents}.
\newblock arXiv:2610.21900v1, 2026.
\newblock URL \url{https://arxiv.org/abs/2610.21900v1}.

\bibitem[Shen(2026)]{shen2026eviction}
C.~Shen.
\newblock {What eviction destroys: A restore-counterfactual audit of forgetting
  in agent memory}.
\newblock arXiv:2609.08279v1, 2026.
\newblock URL \url{https://arxiv.org/abs/2609.08279v1}.

\end{thebibliography}
\end{document}